\documentclass{article}

\usepackage{microtype}
\usepackage{graphicx}
\usepackage{subfigure}
\usepackage{booktabs} 
\usepackage{hyperref}
\usepackage{hyperref}

\usepackage[accepted]{icml2021}

\usepackage{amsmath}
\usepackage{amssymb}
\usepackage{amsthm}
\usepackage{amsfonts}
\usepackage{caption}
\usepackage{float}
\usepackage{multirow}

\usepackage{color}
\definecolor{darkgreen}{rgb}{.1,0.4,.1}

\definecolor{darkred}{rgb}{0.6,0,0}

\icmltitlerunning{Dissipative Hamiltonian Neural Networks}

\begin{document}

\twocolumn[
\icmltitle{Dissipative Hamiltonian Neural Networks:\\ Learning Dissipative and Conservative Dynamics Separately}

\begin{icmlauthorlist}
\icmlauthor{Andrew Sosanya}{mlc} 
\icmlauthor{Sam Greydanus}{mlc}
\end{icmlauthorlist}

\icmlaffiliation{mlc}{The ML Collective} 
\icmlcorrespondingauthor{Andrew Sosanya}{andrewsosanya@gmail.com}

\icmlkeywords{hamiltonians, physics, neural networks, vector fields, Neural ODEs}

\vskip 0.3in
]

\printAffiliationsAndNotice{\icmlEqualContribution} 


\begin{abstract}
    Understanding natural symmetries is key to making sense of our complex and ever-changing world. Recent work has shown that neural networks can learn such symmetries directly from data using Hamiltonian Neural Networks (HNNs). But HNNs struggle when trained on datasets where energy is not conserved. In this paper, we ask whether it is possible to identify and decompose conservative and dissipative dynamics simultaneously. We propose Dissipative Hamiltonian Neural Networks (D-HNNs), which parameterize both a Hamiltonian and a Rayleigh dissipation function. Taken together, they represent an implicit Helmholtz decomposition which can separate dissipative effects such as friction from symmetries such as conservation of energy. We train our model to decompose a damped mass-spring system into its friction and inertial terms and then show that this decomposition can be used to predict dynamics for unseen friction coefficients. Then we apply our model to real world data including a large, noisy ocean current dataset where decomposing the velocity field yields useful scientific insights.
\end{abstract}

\begin{figure*}[ht!]
    \centering
    \includegraphics[width=0.85\textwidth]{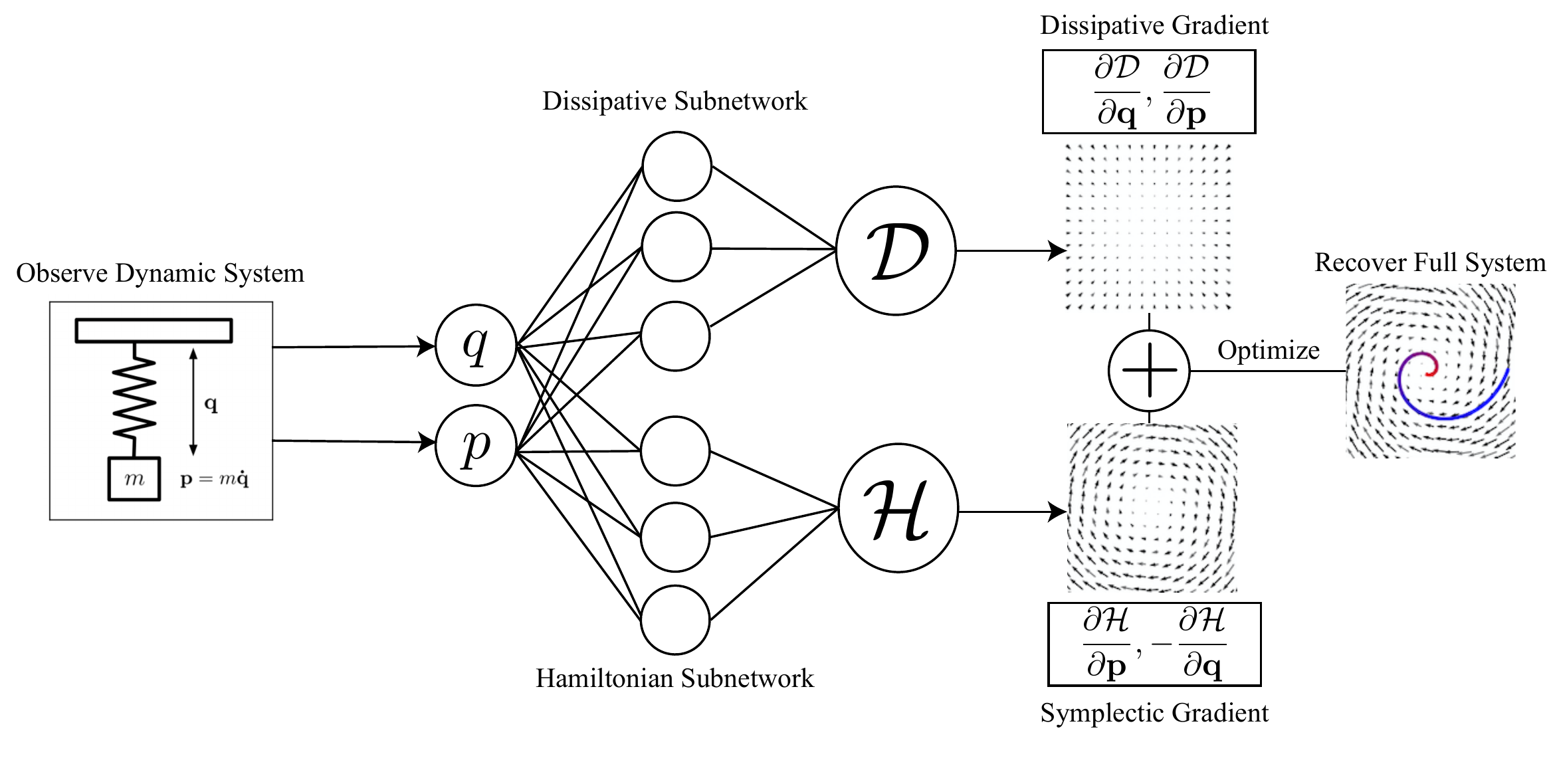}
    \captionof{figure}{Visualizing the architecture of a Dissipative Hamiltonian Neural Network (D-HNN). D-HNNs leverage two neural networks to model dynamic systems. As with a vanilla HNN, we parameterize a Hamiltonian function $\mathcal{H}$ with the first subnetwork and use it to model part of the system's dynamics (lower part of diagram). Unlike the vanilla HNN, we use a second subnetwork to allow our parametric model to output a second scalar function, $\mathcal{D}$, and use it to obtain an ordinary, non-symplectic gradient which can be used to model dissipative dynamics. The sum of these two fields can be used to model dynamic systems which have both conservative and dissipative dynamics: in this case, a damped spring. }
    \label{fig:hero}
\end{figure*}

\section{Introduction}

We are immersed in a complex, dynamic world where change is the only constant. And yet there are certain patterns to this change that suggest natural laws. These laws include conservation of mass, energy, and momentum. Taken together, they constitute a powerful simplifying constraint on reality. Indeed, physics tells us that a small set of laws and their associated invariances are at the heart of all natural phenomena. Whether we are studying weather, ocean currents, earthquakes, or molecular interactions, we should take care to respect these laws. And when we apply learning algorithms to these domains, we should ensure that they, too, take these laws into account.

One of the remarkable insights of physics is that all of these laws can be expressed as conserved quantities. For example, conservation of angular momentum enforces rotational invariance and conservation of mass-energy enforces time-translation invariance \cite{noether1971invariant}. The mathematical framework for representing these conserved quantities is called Hamiltonian mechanics, and the function which relates the coordinates of a physical system to a conserved quantity (usually energy) is called the Hamiltonian. Recent work by \citet{greydanus2019hamiltonian}, \citet{toth2019hamiltonian} and others has shown that this function can be used to incorporate physical constraints, like conservation of energy, into deep neural networks.

These models, which we will refer to as Hamiltonian Neural Networks (HNNs), are able to learn more accurate representations of the world than would otherwise be possible. Recent works by \citet{cranmer2020lagrangian}, \citet{ chen2019symplectic}, and \citet{choudhary2020physics} have shown how to learn Hamiltonians for nonlinear, chaotic systems such as double pendulums, $n$-body systems, and bouncing balls. Moreover, these models can be learned starting from perceptually difficult observations, such as pixel videos. A number of other works have begun applying these physics-inspired models to new fields such as quantum mechanics \cite{broughton2020tensorflow}, chaos theory \cite{choudhary2020physics}, and differential equations \cite{mattheakis2020hamiltonian}.

One remaining drawback of HNNs is that they assume that the total energy of the observed system is conserved. In other words, they assume the absence of friction and other sources or sinks of energy. This obscures an important fact about the real world: no matter how sensitive one's measurements are, real-world data never conforms exactly to theoretical expectations. Real-world dynamic systems often contain dissipative dynamics due to friction, external forces, or measurement noise. How might we enable HNNs to learn conserved quantities even in the presence of noisy real-world data?

In this paper, we tackle the problem of learning in real-world dynamic systems by introducing Dissipative Hamiltonian Neural Networks (D-HNNs), a model that extends HNNs to learn the decomposition of any dynamic system into its dissipative and conserved quantities. D-HNNs learn decompositions by learning a Rayleigh dissipation function $\mathcal{D}$ in addition to a Hamiltonian. The resulting network has two scalar outputs, $\mathcal{H}$ and $\mathcal{D}$. The first yields a rotational vector field with respect to the input coordinates and the second yields an irrotational vector field over the same coordinates. Summed together, the Hodge-Helmholtz theorem tells us that these two vector fields can fit any arbitrary vector field. By the same token, any vector field -- say, a dataset of velocity measurements -- can be decomposed into the sum of a rotational vector field and an irrotational vector field. We call this decomposition the Helmholtz decomposition \cite{bhatia2012helmholtz}.

When we train our model on a dataset, it learns an implicit Helmholtz decomposition of the dynamics. In other words, it learns the dissipative and conservative dynamics separately. When we train it on observations of a damped spring, for example, our model learns a Hamiltonian resembling that of a non-damped spring while using the Rayleigh dissipation function to model the effects of friction. Later in this paper, we show that by multiplying the Rayleigh function by various constants, we can estimate what dynamics would occur for different friction coefficients. Thus we can answer counterfactual questions about a system such as \textit{what would happen if the friction were halved?} On the spring task and others, we show that D-HNNs match or outperform baseline models and non-ML numerical approaches.

Learning decompositions of dynamic systems has practical applications for other fields of vector analysis such as computer graphics, fluid dynamics, weather modeling, and even forensics \cite{bhatia2012helmholtz}. One example we consider in this work is oceanography. Oceanographers are often interested in locating upwelling, a dissipative feature, which occurs when cold, nutrient-rich waters rise to the surface of the ocean, supporting large populations of fish and other marine life. Finding such dissipative patterns in ocean current data is a non-trivial task. In this work, we apply D-HNNs to a large ocean-current dataset and use them to map such features.

\section{Related Work} 

\textbf{Physics for ML.} Recent works have embedded physical priors into learning processes to narrow the effective search space and regularize the training process. For example, \citet{xie2018tempogan} generated realistic fluid flows with coherence by placing physical constraints on the loss term. \citet{cranmer2020lagrangian} and \citet{lutter2019delan} used neural networks to parameterize Lagrangians and thereby enforce the principle of stationary action. \citet{cranmer2020discovering} used a Graph Neural Network and symbolic regression to extract symbolic representations of physical laws from data. Through a physics-as-inverse graphics approach, \citet{jaques2019physics} used a model to perform unsupervised physical parameter estimation of systems starting from video. \citet{Schmidt2009Distilling} used a genetic algorithm to discover conservation laws and recover the symbolic Lagrangians and Hamiltonians of harmonic oscillators and chaotic double pendulums from data.

\textbf{Hamiltonian Neural Networks.} \citet{greydanus2019hamiltonian} introduced the concept of using Hamiltonian's equations as an inductive bias within neural networks to learn conservation laws from data. HNNs learn a parametric function, $\mathcal{H}_\theta (\textbf{q},\textbf{p})$, which functions as the Hamiltonian and maps coordinates to their approximate time derivatives. In order to train an HNN, one minimizes the error between the known time derivatives of coordinates $p$ and $q$ and the symplectic gradient of $\mathcal{H}$ with respect to the input coordinates: $\frac{\partial \mathcal{H}}{\partial \textbf{p}} = \frac{\partial \textbf{q}}{dt},  -\frac{\partial \mathcal{H}}{\partial \textbf{q}} = \frac{\partial \textbf{p}}{dt}$. This structure enables HNNs to learn conservation laws from arbitrary coordinates. \citet{greydanus2019hamiltonian} also combined HNNs with an autoencoder and used them to model the dynamics of a real pendulum using a dataset of pixel observations. HNNs successfully learned conservation laws from ideal and simple systems, but struggled with modeling more complex systems such as the three-body problem. They are also limited to \textit{only} extracting conserved quantities from a given system and thus cannot capture dissipative dynamics in real systems -- a key difference compared to our work.

\textbf{Hamiltonian Generative Networks.} \citet{toth2019hamiltonian} introduced Hamiltonian Generative Networks (HGNs), a class of generative models that can learn time-reversible Hamiltonian dynamics in an abstract phase space, starting from image inputs. HGNs learn through a three-step process. First, they encode a sequence of images into an initial state $s_t$ within abstract state space and map those states to a scalar $\mathcal{H}_\gamma(s_t) \in \mathbb{R}$, which can be interpreted as the Hamiltonian. Next, dynamics are estimated in the abstract state space using the Hamiltonian and a deconvolutional network projects the result back to a pixel representation. Finally, the network is optimized using a loss analogous to that of a temporally-extended variational autoencoder. HGNs excel at learning Hamiltonians from dynamic systems with high-dimensional representations. However, like HNNs, they only extract conserved quantities and cannot model the dissipative components of a real system. 

\textbf{Other Hamiltonian-based models.} Other works have also built on the foundations of HNNs. Symplectic Recurrent Neural Networks by \citet{chen2019symplectic} learn better Hamiltonians than HNNs using a symplectic integrator, multi-step training and initial state optimization.  \citet{sanchez2019hamiltonian} combine graph networks with differentiable ODE integrators and Hamiltonian inductive biases to predict the dynamics of particle systems. \citet{zhong2019symplectic} combine external control methods with the Hamiltonian formalism to perform continuous control and planning with the learned model. Concurrent to our work, \citet{zhong2020dissipative} trained a Neural ODE to learn the dynamics of dissipative systems. However, their model does not model dissipative quantities separately and it does not offer advantages in accuracy over baseline models.

\begin{figure*}
    \begin{center}
      \includegraphics[width=.9\textwidth]{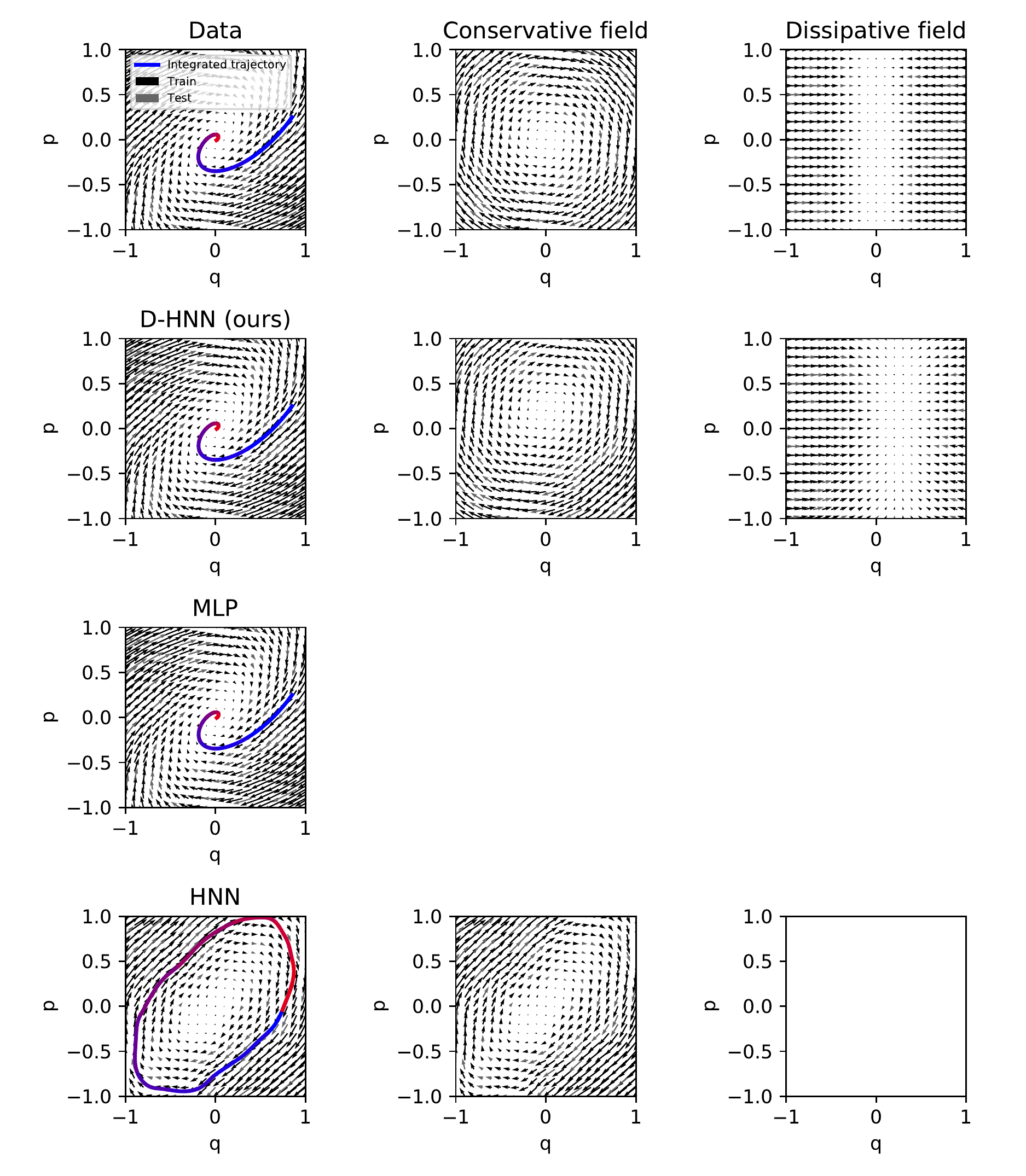}
        \captionof{figure}{Training a Dissipative Hamiltonian Neural Network (D-HNN) and several baseline models on a damped spring task. \textbf{Row 1.} We decompose the original dataset by interpolating the training data onto a grid with a nearest-neighbors approach and then decomposing the field into two components. In practice, we accomplish this using a few hundred iterations of the Gauss-Seidel method to solve Poisson's equation. \textbf{Row 2.} We train a D-HNN on the same data. The first half of this model looks exactly like an HNN. The second half also looks like an HNN, except we use the gradient of the scalar field directly rather than rearranging it according to Hamilton's equations. This gives us a trainable \textit{dissipative} field which is able to model the dissipative components of the damped spring's dynamics. Summing these two fields, we obtain an accurate model of the dynamics of the system. \textbf{Row 3.} We train the baseline model (MLP) on the training set; this model gives a good fit and can be integrated as a Neural ODE, but it cannot be used to decompose the field into conservative and dissipative components. \textbf{Row 4.} We train a Hamiltonian Neural Network (HNN) on the same dataset and find that it is only able to model the conservative component of the system's dynamics. In other words, it strictly enforces conservation of energy in a scenario where energy is not actually conserved.}
    \label{fig:dampedspring}
    \end{center}
\end{figure*}

\section{Theory}

D-HNNs learn to decompose the dynamics of physical systems into their dissipative and conservative components. To do this, they employ the tools of Hamiltonian mechanics and the Helmholtz decomposition.

\subsection{Hamiltonian mechanics}


We begin by providing a brief overview of Hamiltonian mechanics, an essential tool for modeling the time evolution of dynamic systems. The Hamiltonian $\mathcal{H}(\textbf{q},\textbf{p})$ is scalar function such that
\begin{equation}
   \frac{\partial \mathcal{H}}{\partial \textbf{p}} = \frac{\partial \textbf{q}}{dt},  -\frac{\partial \mathcal{H}}{\partial \textbf{q}} = \frac{\partial \textbf{p}}{dt}
\end{equation}
where $(\textbf{q},\textbf{p})$, composed of pairs $(q_1, p_1)...(q_N, p_N)$, are the coordinates of the system.  Moving in the direction of the \textit{symplectic} gradient $ S_\mathcal{H} = (\frac{\partial \mathcal{H}}{\partial \textbf{p}}, -\frac{\partial \mathcal{H}}{\partial \textbf{q}})$ keeps the value of $\mathcal{H}$'s output constant throughout the time evolution of the system, thereby representing a time-independent conserved quantity. Using $\mathcal{H}$, we can model the exact conserved quantities of both simple and complex dynamic systems. 

\textbf{Nonconservative systems.} Real dynamic systems often contain dissipative forces such as friction or radiation, and these forces cause energy loss over time. Thus energy is not strictly conserved, violating the assumptions of regular Hamiltonian mechanics. Therefore we must work within the framework of nonconservative systems when predicting dynamics in dissipative environments. The Rayleigh dissipation function $\mathcal{D}(\dot{\textbf{q}})$ is a scalar function that provides a way to account for generalized, velocity-dependent dissipative forces—such as friction—in the context of Hamiltonian mechanics \cite{cline2017variational}. Equation \ref{eq:lin-Dissipative} shows the Rayleigh function for linear, velocity-dependent dissipation. Here $\rho$ is a constant and $\dot q$ is the velocity coordinate.
\begin{equation}
    \mathcal{D} = \frac{1}{2}\rho\dot{q}^2
    \label{eq:lin-Dissipative}
\end{equation}
Equation \ref{eq:nonlin-Dissipative} shows that nonlinear, velocity-dependent dissipation forces $F_i$ within the system can be accounted for as long as they are defined as 
\begin{equation}
    F^{f}_i = -\frac{\partial \mathcal{D}(\dot{\textbf{q}})}{\partial \dot{\textbf{q}}}
    \label{eq:nonlin-Dissipative}
\end{equation}
In-depth applications of the Rayleigh dissipation function can be found in \cite{minguzzi2015rayleigh}. 

\subsection{Helmholtz decomposition}
Like many students today, Hermann von Helmholtz realized that medicine was not his true calling. Luckily for us, he pursued physics and discovered one of the most useful tools in vector analysis: the Helmholtz decomposition. Helmholtz's theorem states that any smooth vector field $V$ can be expressed as a sum of its irrotational component $V_{irr}$ and its rotational component $V_{rot}$ \cite{helmholtz1858integrale}. More formally, $V$ can be expressed as the sum of the curl-less gradient of a scalar potential $\phi$ and the divergence-less curl of a vector potential $\mathcal{\textbf{A}}$.
\begin{equation}
    V = V_{irr} + V_{rot} = \nabla\phi + \nabla\times \mathcal{\textbf{A}}
\end{equation}
D-HNNs performing an implicit Helmholtz decomposition can be found in Figure \ref{fig:dampedspring} and Figure \ref{fig:dampedspring_counterfactual}.

\textbf{Applications.} Scientists have leveraged the Helmholtz decomposition to uncover features of interest in their data or to model different phenomena, most commonly different types of fluids. For example, the Helmholtz decomposition has been used to match fingerprints \cite{palit2005applications}, model geomagnetic fields \cite{akram2010regularisation}, and accurately animate fire \cite{nguyen2002physically}.

\textbf{Approximations.} The Helmholtz decomposition can be implemented with approximate numerical methods such as the least squares finite-element method \cite{bhatia2012helmholtz, stam2003real, bell1991efficient}. While the decomposition of a vector field is uniquely determined with appropriate boundary conditions \cite{griffiths2005introduction}, numerical decompositions are not: just as there are numerous ways for a model to fit data, there are numerous ways to decompose a vector field. In this work, for example, we follow \citet{stam2003real}, using the Gauss-Seidel method to obtain baseline numerical decompositions. One problem with most of these methods is that they are iterative and not easily parallelizable on GPUs. When dealing with a large dataset, the \textit{entire} dataset typically needs to be considered, which can be computationally expensive and memory-intensive.

There is an even more insidious source of error: the grid-based interpolation one must do prior to decomposition. In our experiments, for example, the error due to imperfect interpolation was many orders of magnitude larger than the error due to numerical decomposition. In contrast, neural networks excel at interpolating data \cite{hornik1989multilayer}. We suspect that this is one reason they were able to improve over baseline approaches.

\section{Methods}

D-HNNs perform an implicit Helmholtz decomposition on a dynamic system's vector field $V$ that describes all the possible states of the system—and extract $V_{irr}$ and $V_{rot}$, which correspond to the dissipative and conserved quantities of the system, respectively. D-HNNs are parameterized as a neural network, with parameters $\theta$, that leverage two subnetworks to learn two differentiable scalar functions, a dissipative function $\mathcal{D}(\textbf{q},\textbf{p})$ and a Hamiltonian function $\mathcal{H}(\textbf{q},\textbf{p})$. During the forward pass, each subnetwork digests input canonical coordinates (here $\textbf{q},\textbf{p} \in \mathbb{R}^n$) that describe the state of the system and then outputs scalar values for $\mathcal{D}$ and $\mathcal{H}$. Next, we take the in-graph gradient of both scalars with respect to the inputs. Finally, we compute an L2 loss between the sum of these input gradients and the target gradients, and backpropagate error to all trainable parameters. Equation \ref{eq:loss} gives the particular form of our loss function.
\begin{equation}
    \begin{split}
    L_{DHNN} = \bigg \Vert (\frac{\partial \mathcal{H}}{\partial \textbf{p}} + \frac{\partial \mathcal{D}}{\partial \textbf{q}}) - \frac{\partial \textbf{q}}{\partial t} \bigg \Vert_2 \\
    + \bigg \Vert (-\frac{\partial \mathcal{H}}{\partial \textbf{q}} +  \frac{\partial \mathcal{D}}{\partial \textbf{p}}) - \frac{\partial \textbf{p}}{\partial t} \bigg \Vert_2
    \end{split}
    \label{eq:loss}
\end{equation}
Intuitively, we begin by summing the symplectic gradient of $\mathcal{H}$ and the regular gradient of $\mathcal{D}$. This quantity should match the time derivatives of the input coordinates after training. In order to make this happen, we compute an L2 loss between it and the true time derivatives of the system. Figure \ref{fig:hero} provides a visual summary of this process. 

\begin{figure}[ht!]
\centering
\includegraphics[width=.5\textwidth]{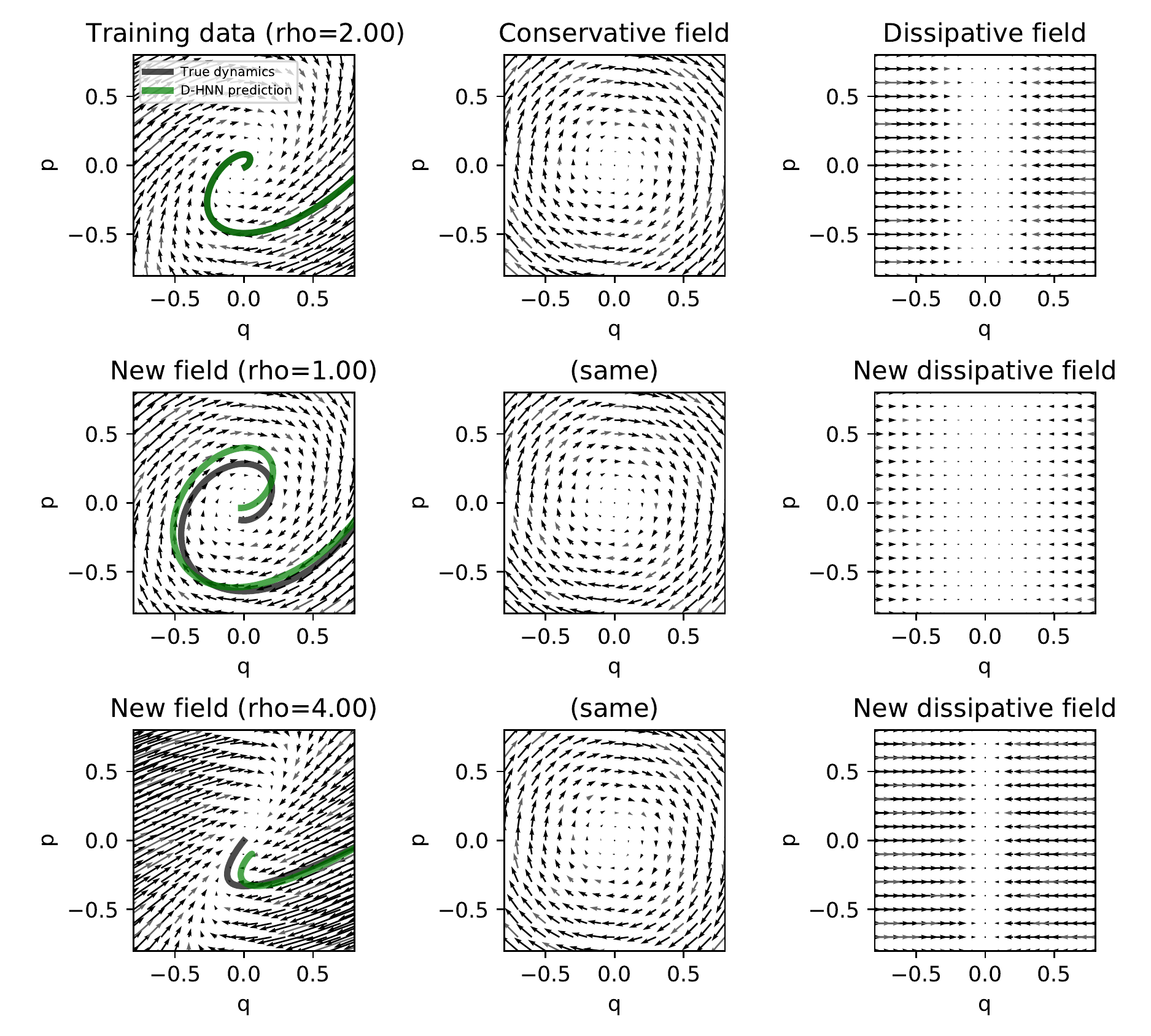}
\caption{Adjusting the dissipative component of a trained D-HNN so as to predict trajectories for \textit{unseen} friction coefficients. Adjustments to the coefficients can be made by multiplying the learned Rayleigh dissipation function by various constants.}
\label{fig:dampedspring_counterfactual}
\end{figure}

Because D-HNNs learn exact decompositions, they can recover the dynamics of the full system, which is simply the sum of the dissipative and conserved dynamics. These decompositions even permit new forms of generalization. For example, we can multiply the dissipative component by various coefficients in order to answer the counterfactual question, \textit{what would the dynamics look like if we halved or doubled the friction coefficient?} Figure \ref{fig:dampedspring_counterfactual} does this with the damped spring model; we see that the D-HNN's trajectories are close to ground truth even though it was never exposed to different friction coefficients during training. We emphasize that other parametric models such as MLPs are unable to perform this kind of generalization.

D-HNNs can also learn time-dependent Hamiltonians and Rayleigh dissipation functions. This is done by including $t$ as an auxiliary input variable: $ (\textbf{q}, \textbf{p}, t)$. We do not take any gradients with respect to $t$. This capability is important for Task 3, where the system's dynamics are highly time-dependent. 

\textbf{Tasks.} We evaluate D-HNNs by testing their ability to decompose three dynamic systems: a damped mass-spring oscillator, a real damped pendulum, and surface currents taken from satellite data. The first two tasks are almost identical to the original HNN experiments in \cite{greydanus2019hamiltonian}, but they differ in that here we intentionally model dissipative dynamics in addition to conservative dynamics. Since the Hamiltonians of these systems are invariant with respect to time, we set $t = 0$ in all cases. 

\textbf{Baselines.} We compare the performance of D-HNNs to HNNs, a baseline multi-layer perceptron (MLP) model, and a numerical decomposition method because they have the most relevance to our performance metrics.


\begin{table*}[ht!]
\caption{Comparing D-HNN training and evaluation metrics to a number of baseline approaches. The test MSE metric measures the difference between the predicted time derivatives and true time derivatives of the system, and the energy MSE metric measures the differences between the predicted energies and true energies. The D-HNN outperforms the MLP and HNN on all tasks, though sometimes only by a small margin. The MLP is slightly behind the D-HNN in predicting energies and state trajectories. We also tried running a numerical Helmholtz decomposition on the mass-spring dataset (using the Gauss-Seidel method to solve Poisson's equation). This introduced a mean square error of $7.3\cdot 10^{-3}$ which was mostly due to interpolation error.}
\label{tab:main-benchmark}
\begin{center} \begin{small}
\begin{tabular}{lcccc}
\toprule
Metric & Model & Damped spring & Real pendulum & OSCAR ocean data\\
\midrule
\multirow{3}{*}{Test MSE} & MLP & $9.9\cdot 10^{-5}$ & $1.1\cdot 10^{-3}$ & $0.181$ \\
& HNN & $6.3\cdot 10^{-1}$ & $1.4\cdot 10^{-3}$ & $0.146$ \\
& D-HNN & $\mathbf{5.8\cdot 10^{-5}}$ & $\mathbf{9.5\cdot 10^{-4}}$ & $\mathbf{0.143}$ \\
\midrule
\multirow{3}{*}{Energy MSE} & MLP & $6.0\cdot 10^{-6}$ & $1.2\cdot 10^{-3}$ & $-$ \\
& HNN & $2.9\cdot 10^{-1}$ & $1.0\cdot 10^{-2}$ & $-$ \\
& D-HNN & $\mathbf{3.6\cdot 10^{-6}}$ & $\mathbf{1.1\cdot 10^{-3}}$ & $-$ \\
\bottomrule
\end{tabular}
\end{small} \end{center}
\end{table*}

\begin{figure*}[ht!]
\centering
\includegraphics[width=\textwidth]{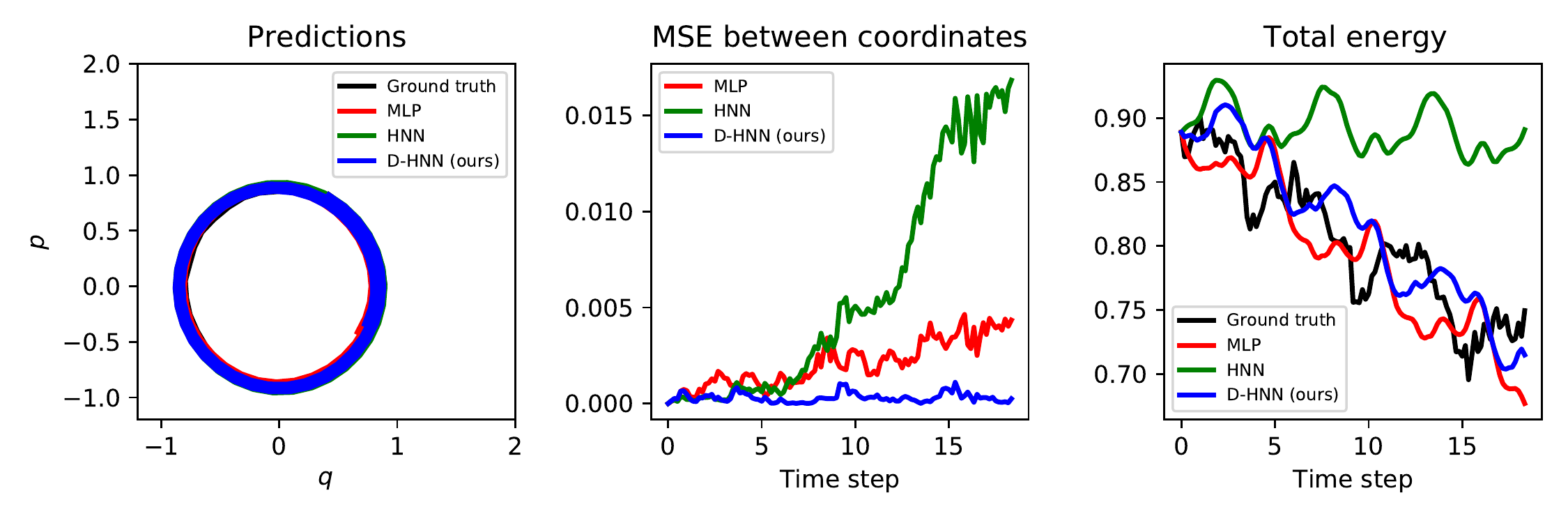}
\caption{Analyzing a model trained on data from a real, slightly damped pendulum. \textbf{Left:} The predicted dynamics from all the models compared to the ground truth. The D-HNN trajectory covers up the MLP and HNN trajectories. The D-HNN recovers the system with high accuracy and obscures the ground truth dynamics. See Figure 6 
for individual trajectories. \textbf{Middle:} The mean square error (MSE) between predicted coordinates and the true coordinates. D-HNNs clearly outperform the other models in providing the best accuracy along a given trajectory. \textbf{Right:} The predicted energies of the pendulum system from each model compared to the ground truth. Here, the D-HNN slightly outperforms the MLP. Thus the D-HNN excels at both predicting the right trajectories and modeling the system's dissipation over time.}
\label{fig:realpend_energy}
\end{figure*}

\textbf{Task 1: Damped spring.} A damped mass-spring system can be described as simple harmonic oscillator with a velocity-dependent frictional force. The Hamiltonian and Rayleigh functions of such a spring can be described by \ref{eq:massspringH} and \ref{eq:lin-Dissipative} respectively. 
\begin{equation}
    \mathcal{H} = \frac{1}{2}kq^2 + \frac{p^2}{2m}
    \label{eq:massspringH}
\end{equation}
Here $k$ is the spring constant, $q$ is the distance from the origin, $\rho$ is the friction coefficient, $p$ is the momentum variable and $m$ is the mass constant. For simplicity, we set $k = m = 1$. We initially set $\rho = 2$, but $\rho = 1, 4$ when making counterfactual adjustments (Figure \ref{fig:dampedspring_counterfactual} to the friction coefficient.

\textbf{Training.} We trained three models on the dataset: a D-HNN, an HNN, and an MLP that directly mapped input coordinates ($\textbf{q},\textbf{p}$) directly to ($ \frac{\partial \textbf{q}}{\partial t}$, $ \frac{\partial \textbf{p}}{\partial t})$. The HNN and MLP models contained two layers of 256 neurons. Each of the D-HNN's subnetworks contained one layer of 256 neurons and a residual layer. We shuffled the dataset prior to training to eliminate order effects. We backpropogated our model using Equation \ref{eq:loss} as our loss function. We synthetically constructed the dataset and analytically calculated the targets using Equation \ref{eq:massspringH} and \ref{eq:lin-Dissipative}. We used the analytical time derivatives of the system as the targets. On each task, we trained each model with a learning rate of $10^{-2}$ for 5000 gradient steps, using a 80/20 train/test percentage split (see Appendix for more on hyperparameters). We used the Adam optimizer \cite{Kingma2014Adam} and conducted batch gradient descent with a batch size of 128. We implemented our models in PyTorch and trained each model with P100 GPUs. 

\textbf{Metrics.} At inference time, we report the mean-squared error (MSE) between the true coordinates and the target coordinates. To obtain the reported dynamics and energy metrics, we used an initial state starting at $(0.9, 0)$ and used the fifth-order Runge-Kutta integrator in \texttt{scipy.integrate.solve\_ivp} with the error tolerance set to $10^{-12}$. We also compared our models' decomposition performance to the Gauss-Seidel method, an iterative numerical decomposition that solves Poisson's Equation (for more details, see \cite{stam2003real}).

\textbf{Task 2: Real Damped Pendulum System.} A damped pendulum system can be described as a nonlinear harmonic oscillator inhibited by a frictional force. The Hamiltonian and Rayleigh functions of such a system can be respectively described by Equation \ref{eq:pendulum} and \ref{eq:lin-Dissipative} respectively. 
\begin{equation}
    \mathcal{H} = 2mgl(1-\cos{q}) + \frac{l^2p^2}{2m}
    \label{eq:pendulum}
\end{equation}
For simplicity, we set $l = m = 1$. However, we set our gravity constant $g = 3$. We perform the same training procedures and metrics reporting that were done Task 1, with three exceptions. First, we feature position and momentum readings of a real pendulum with friction from \citet{Schmidt2009Distilling}. Second, we do not use the Gauss-Seidel decomposition as a comparison because the dataset already consists of trajectories, and thus we do not need to use an analytic method to estimate a trajectory. Third, we report the MSE of the total and predicted energy by integrating along the trajectory from the first point in the test set.

\textbf{Task 3: Surface currents.} This is perhaps the most interesting task: we test D-HNNs on decomposing velocity fields of surface currents. We use the OSCAR dataset from Earth and Space Research (See Appendix). For our task, we extracted 69 time frames, each consisting of a 50x50 grid of surface currents from the Atlantic Ocean (see Appendix). Each point in the grid had longitudinal and latitudinal coordinates $(x,y)$ as well as zonal and meridional current velocities $(u,v)$ measured in five-day intervals. On this task, we used the same methods described in Task 1, except for three differences. First, we trained for 24000 steps to handle the large amount of data (69x50x50 points). Second, we normalized the input coordinates into the range $[-1,1]$ in the preprocessing stage because of the higher variances within the range of the target velocities in the dataset. Third, we fed in time as an auxiliary variable into our model. We normalized our time variable $t$ into the range $[-1,1]$.

\section{Results}

\textbf{Damped mass-spring results.} Figure \ref{fig:dampedspring} shows the results of  training the baseline model (MLP), a HNN and a D-HNN on a damped mass-spring dataset. The D-HNN constructed an exact decomposition of the system, extracted its conservative and dissipative quantities, and modeled its dynamics with high-accuracy. It outperformed both the HNN and MLP as well as the Gauss-Seidel numerical decomposition. The MLP gives a good fit and can be integrated as a Neural ODE \cite{chen2018neural}, but it cannot reproduce a decomposition of the the dissipative and conservative quantities of the system. The HNN was only able to extract the conserved quantity of the mass-spring system when trained with the same dataset. The HNN yielded the worst fit of the three models because it incorrectly assumed that energy is strictly conserved.  

Because D-HNNs can perform decompositions, we were able to test their ability to answer counterfactual questions related to the dissipative component, for example, \textit{How does the friction coefficient affect the dynamics of the mass-spring system?} As seen in Figure \ref{fig:dampedspring_counterfactual}, D-HNNs can predict the trajectories in systems with different friction coefficients. We emphasize that the D-HNN model only saw dynamics corresponding to one friction coefficient during training. We note that, because D-HNNs are an extension of HNNs, they can also predict trajectories with unseen modifications of the conservative components, as seen in Figure 5 of \citet{greydanus2019hamiltonian}.

\textbf{Real damped pendulum results.} 
Figure \ref{fig:realpend_energy} shows that the D-HNN improves over the baseline model and the HNN in terms of both mean square error between coordinate trajectories and mean square error between predicted and actual total energy. The baseline MLP, meanwhile, does a slighly worse job of predicting a trajectory and conserving total energy. However, it still improves over the HNN. The HNN accrues the most error out of the three models, by a factor of ten in both trajectory and energy MSE. We observe this effect because HNNs explicitly force energy to be conserved in the system, which is poor assumption for this dataset, which contains some friction.

\textbf{Surface current results.} The ways of the ocean are complex. Oceanographers are typically interested in modeling ocean flow and identifying features such as eddy currents, upwelling and downwelling. \cite{zhang2019helmholtz, chelton2011global, petersen2013three}. Upwelling is a dissipative feature that occurs when cold, nutrient-rich waters rise to the surface of the ocean. Conversely, downwelling is an opposite process by which surface water descends. These features tend to occur in the centers of vortices, which form in order to preserve the geostrophic balance. Oceanographers often describe current velocity fields by composing a divergence-free stream function $\psi$ (conserved) and some curl-less velocity potential $\phi$ (dissipative) \cite{buhler2014wave, caballero2020integration}. 

\begin{figure}[ht!]
\centering
\includegraphics[width=.49\textwidth]{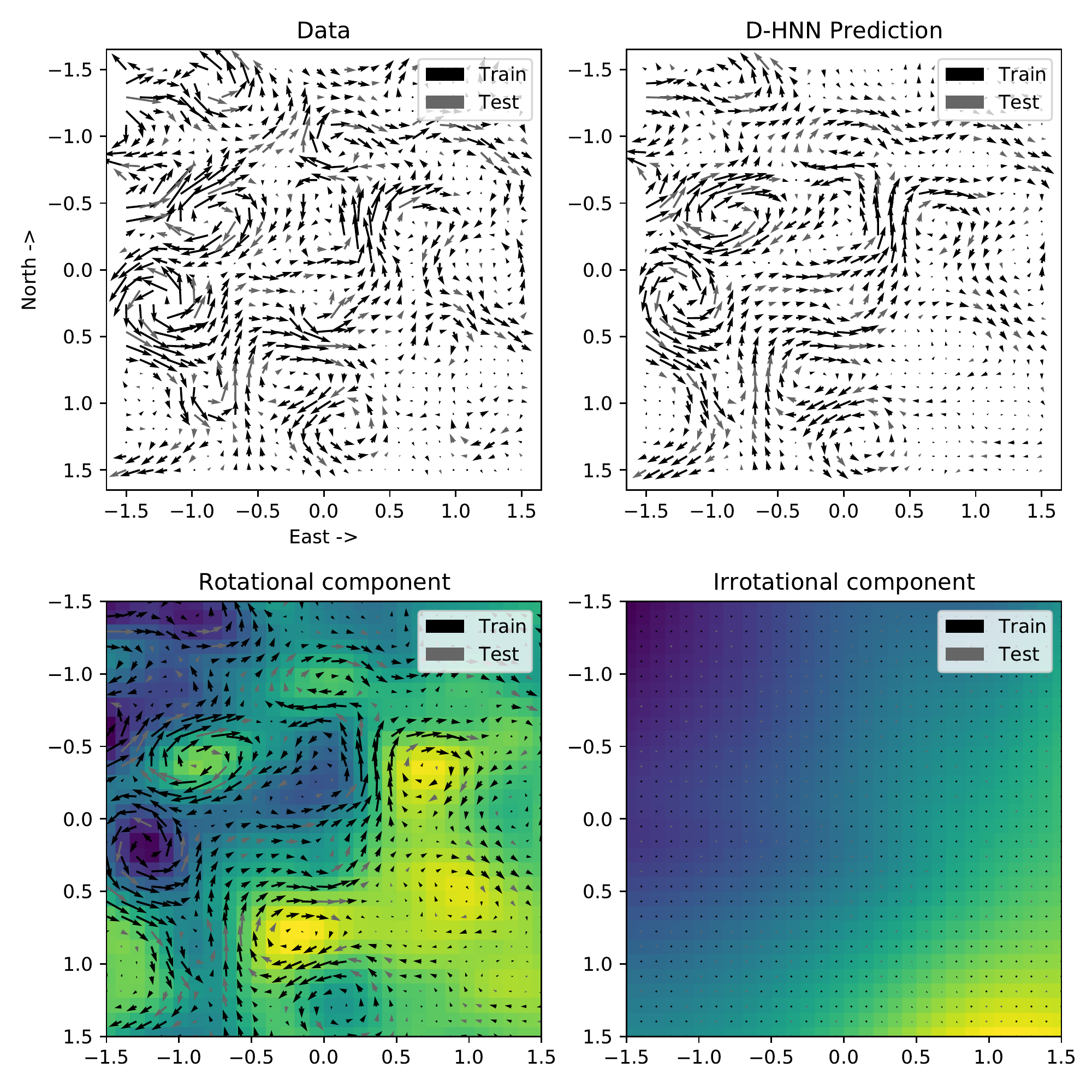}
\caption{A learned decomposition of surface current velocities from the OSCAR ocean dataset. Backgrounds in plots 3 and 4 are colored according to scalar fields $\mathcal{H}$ and $\mathcal{D}$ respectively. In Plot 3, the regions which are particularly blue or yellow correspond to centers of vortices. These regions tend to be associated with upwelling and downwelling and are of ecological importance.}
\label{fig:ocean_vfield}
\end{figure}

Our experiment showed that D-HNNs can separately model both $\psi$ and $\phi$—both unknown \textit{a priori}—and uncover ocean flow features by performing an implicit Helmholtz decomposition across 69 time frames. From the bright and dark regions seen in the $H$'s learned scalar fields (Figure \ref{fig:ocean_vfield}, we can identify eddies by locating the centers of vortices. To maintain geostrophic balance in a vortex, water will move upwards or downwards depending on the direction of the vortex. By identifying the centers of vortices, we can find where upwelling and downwelling features will be the strongest.

D-HNNs only performed decompositions slightly better than HNNs because ocean flow is typically much more conservative than dissipative and thus makes modeling an easier task for HNNs. Eddies have a much more prominent effect on the velocity fields than the dissipative features of upwelling and downwelling (Figure \ref{fig:ocean_vfield}). We find that upwelling and downwelling does occur within the examined region but with a much fainter effect on the velocity field than eddies. This same effect is observed in \citet{zhang2019helmholtz} where the authors deploy non-ML Helmholtz decompositions for eddy detection. We would like to investigate this in future experiments. 

Additionally, we found that D-HNNs are able to upsample decompositions across the time domain because the model's outputs can be sampled at higher frame rates than the original training data. Thus one can easily interpolate between two trames of training data. For example, for frames 10 and 11, we can approximate the decomposition of frame 10.5. By this means, we can use D-HNNs to, say, double video frame rates.

\section{Discussion}
D-HNNs build on the foundations of HNNs by bridging the gap between the clean, natural symmetries of physics and the imperfect, messy data often found in the real world. They employ the tools of Hamiltonian mechanics and Helmholtz decomposition to separate conserved quantities from dissipative quantities. In doing so, they allow us to model complex physical systems while enforcing strict conservation laws. They represent a small, practical advance in that they can help us see our datasets, like the ocean currents dataset, in a new way. And at the same time, they represent progress towards the more ambitious goal of building models which make sense of the incredible complexity of the real world by focusing on its invariant quantities.

\section*{Acknowledgements}

The authors would like to thank Miles Cranmer, Stephan Hoyer, Jason Yosinski, and Misko Dzamba for prior collaborations and fruitful conversations that helped lay the groundwork for this line of inquiry. The authors would also like to thank Robyn Millan, Kristina Lynch, and Mathias Van Regemortel for their encouragement in this intellectual pursuit. 

\bibliographystyle{abbrvnat}
\setlength{\bibsep}{5pt} 
\setlength{\bibhang}{0pt}
\bibliography{references}


\end{document}